\documentclass[twocolumn, switch]{article} 

\usepackage{preprint}

\usepackage{amsmath, amsthm, amssymb, amsfonts}
\usepackage{algorithm}
\usepackage{algorithmic}

\usepackage[numbers,square]{natbib}

\usepackage[utf8]{inputenc}	
\usepackage[T1]{fontenc}	
\usepackage{xcolor}		
\usepackage[colorlinks = true,
            linkcolor = purple,
            urlcolor  = blue,
            citecolor = cyan,
            anchorcolor = black]{hyperref}	
\usepackage{booktabs} 		
\usepackage{nicefrac}		
\usepackage{microtype}		
\usepackage{lineno}		
\usepackage{float}			
\usepackage{hyperref}
\usepackage{cleveref}

\usepackage{dblfloatfix}

\usepackage{lipsum}		

\usepackage{newfloat}
\DeclareFloatingEnvironment[name={Supplementary Figure}]{suppfigure}
\usepackage{sidecap}
\sidecaptionvpos{figure}{c}

\usepackage{titlesec}
\titlespacing\section{0pt}{12pt plus 3pt minus 3pt}{1pt plus 1pt minus 1pt}
\titlespacing\subsection{0pt}{10pt plus 3pt minus 3pt}{1pt plus 1pt minus 1pt}
\titlespacing\subsubsection{0pt}{8pt plus 3pt minus 3pt}{1pt plus 1pt minus 1pt}

\usepackage{tikz,xcolor,hyperref}

\definecolor{lime}{HTML}{A6CE39}
\DeclareRobustCommand{\orcidicon}{
	\begin{tikzpicture}
	\draw[lime, fill=lime] (0,0)
	circle [radius=0.16]
	node[white] {{\fontfamily{qag}\selectfont \tiny ID}};
	\draw[white, fill=white] (-0.0625,0.095)
	circle [radius=0.007];
	\end{tikzpicture}
	\hspace{-2mm}
}
\foreach \x in {A, ..., Z}{\expandafter\xdef\csname orcid\x\endcsname{\noexpand\href{https://orcid.org/\csname orcidauthor\x\endcsname}
			{\noexpand\orcidicon}}
}

\title{GenTrack3: Hybrid Stochastic–Deterministic Online Multi-Object Tracking with Cluster-Aware Association}

\usepackage{xwatermark}
\newwatermark[firstpage,color=gray!90,angle=0,scale=0.28, xpos=0in,ypos=-5in]{*correspondence: \texttt{toanvn@mmmi.sdu.dk}}

\usepackage{authblk}

\author[1]{Toan Van Nguyen\orcidA{}}
\author[2]{Rasmus G. K. Christiansen\orcidB{}}
\author[3]{Dirk Kraft\orcidC{}}
\author[4]{Leon Bodenhagen\orcidD{}}

\affil[*]{SDU Robotics, University of Southern Denmark}

\begin{document}

\twocolumn[ 
  \begin{@twocolumnfalse} 

\maketitle

\begin{abstract}

Multi-object tracking (MOT) involves maintaining consistent target identities as objects dynamically enter and leave a scene. Deterministic approaches, such as tracking-by-detection with data association, produce reproducible results and are computationally efficient, but they rely heavily on motion models and are sensitive to noisy detections that can lead to association errors. In contrast, stochastic methods explicitly model uncertainty and can better handle complex non-linear dynamics, albeit at the cost of increased computational complexity and variability arising from random sampling. This paper presents an online MOT framework that integrates deterministic and stochastic principles to achieve robust tracking under uncertainty. Furthermore, a novel track-to-detection matching approach is introduced to enhance scalability with increasing target numbers while supporting group tracking. The tracking inference mechanism employs a tracklet that includes identifiers, states, velocities, track penalties and track ages of targets, supporting a systematic tracking pipeline. Each target is associated with a stochastic particle set to compute the matching cost to detections. Reference implementations of the proposed approach and baseline trackers can be found on GitHub: \href{https://github.com/SDU-VelKoTek/GenTrack3}{\url{https://github.com/SDU-VelKoTek/GenTrack3}}.

\end{abstract}
\vspace{0.35cm}

  \end{@twocolumnfalse} 
] 



\section{Introduction}
\label{sec:intro}

Multi-object tracking (MOT) remains a highly challenging task, primarily due to the presence of factors such as occlusions, non-linear motion patterns, and high inter-object similarity. When objects exhibit strong visual resemblance, tracking algorithms often face difficulties in maintaining consistent identities over time, leading to potential identity switches. Similarly, partial or complete occlusions, caused by environmental obstacles or interactions between objects, can result in temporary or prolonged loss of target trajectories, further complicating the tracking process. In addition, irregular or unpredictable motion dynamics, which frequently occur in real-world scenarios, impose significant challenges on the accurate temporal association of objects across consecutive frames. Effectively addressing these issues is critical for the development of robust and precise MOT systems capable of functioning reliably in practical scenarios.

\subsection{Related works}

Recent advancements in object detection have solidified tracking-by-detection as the dominant strategy in state-of-the-art MOT frameworks, especially in scenarios involving a variable number of targets since frame-level detections facilitate the automatic initialization and termination of tracks while reducing drift accumulation. Within detection-based paradigm, the Kalman filter is extensively employed for motion state updates and position estimation, often in conjunction with tracklet-detection association mechanisms. In \cite{csaba2010multi}, a data-oriented tracking framework reconstructs trajectories of an unknown number of interacting objects via a three-level association. Local analysis generates trajectory segments, constrained merging forms fragments, and global association produces final estimates using the Hungarian algorithm, which also incorporates sub-threshold detections for ambiguous cases. In \cite{zhen2012improving}, multi-person tracking in semi-crowded scenes is formulated as a nonlinear global optimization integrating social grouping behavior into affinity models. The Lagrange dual is solved via a two-stage algorithm using the Hungarian method and K-means clustering. In \cite{bewley2016simple}, SORT tracker provides a practical solution for real-time multi-object tracking, although its effectiveness is strongly dependent on the accuracy of object detections. Subsequently, DeepSORT, as proposed in \cite{wojke2017simple}, enhances the original SORT by incorporating appearance features to mitigate identity switches. In \cite{yifu2022bytetrack}, ByteTrack further advances tracking accuracy by separately managing high and low confidence detections during data association. In \cite{nir2022botsort}, BoT-SORT achieves improved performance through the integration of a refined Kalman filter, camera motion compensation, and appearance cues. In \cite{jinkun2023OCSORT}, OC-SORT addresses occlusion-related noise by generating virtual trajectories based on observations. In \cite{yu2024smiletrack}, SMILEtrack extends ByteTrack and BoT-SORT by incorporating an object detector with a Siamese similarity module to improve appearance-based matching. In \cite{hyeonchul2024ConfTrack}, ConfTrack enhances a Kalman filter-based tracking framework by applying low-confidence penalization and cascading techniques to manage noisy detections. Built upon SORT, BoostTrack, introduced in \cite{vukas2024inboostrack}, proposes a confidence-weighted similarity metric to prioritize reliable detection-tracklet associations in single-stage tracking. Additionally, ambiguities from Intersection over Union (IoU) are reduced using Mahalanobis distance and shape similarity. Confidence refinement allows effective utilization of low-score detections, while interpolation and camera motion compensation maintain benchmark-level accuracy with real-time performance.  

Real-world scenarios often involve non-linear motion and non-Gaussian noise, but Kalman filter-based trackers rely on the assumptions of linear dynamics and Gaussian noise. In contrast,  particle filters can effectively address these challenges, and many methods have been developed to improve tracking accuracy and robustness of particle-based single-object trackers. However, applying particle filters to multi-object tracking presents difficulties due to the high-dimensional state space that increases with the number of objects. Additionally, exploring particle filters for scenarios with unknown and time-varying numbers of targets remains challenging \cite{isard2001bramble, zia2005MCMC, michael2009robust, ming2009detection, kenji2024boosted}. In \cite{isard2001bramble}, a method is presented to address this issue but exhibited poor scalability as the number of targets grew. Thereafter, an approach is introduced in \cite{zia2005MCMC} which employs a modified Metropolis-Hastings algorithm incorporating add-delete and stay-leave operations; however, it was prone to duplicate tracks and decreased reliability when targets changed frequently. These particle-based approaches primarily focus on managing time-varying target counts, giving less consideration to factors such as object scale or appearance. By contrast, detection-based methods in \cite{michael2009robust, ming2009detection, kenji2024boosted} manage the addition and removal of targets more deterministically and place greater emphasis on robust observation models, though they often overlook the optimization of the tracking inference process. Existing online MOT methods remain limited in their ability to simultaneously handle nonlinear dynamics, non-Gaussian noise, and object identity preservation in scenarios with time-varying numbers of targets. To address this gap, GenTrack, introduced in \cite{toan2025pami}, integrates stochastic and deterministic paradigms for multi-object tracking. Unlike conventional particle sampling methods, each target in GenTrack is represented by a particle set that estimates its optimal state without full posterior sampling, while track initiation and removal are determined through deterministic matching history. Particles are generated via Markov Chain Monte Carlo (MCMC) and refined using Particle Swarm Optimization (PSO) prior to track-detection association. Track-detection matching costs and PSO fitness are formulated based on motion consistency, appearance similarity, target interactions, detection confidence, and track penalties, forming a unified tracking framework. Thereafter, GenTrack2, an extension of GenTrack is presented in \cite{toan2025icra} to better manage occlusions and maintain weak track continuity through a novel weak-track updating scheme and velocity regression. However, scalability remains constrained by the global cost matrix of all current tracks and detections used for matching. Furthermore, while the neighbour-based refinement of weak tracks performs well in sparse scenes, it can lead to ambiguities in crowded environments where weak tracks are surrounded by numerous others.                  


\subsection{Contributions and organizations}

This paper introduces an online MOT framework that integrates deterministic and stochastic principles to enable robust and scalable tracking under uncertainty. The proposed framework, referred to as GenTrack3, models each target through tracklets encoding identifiers, states, velocities, penalties, and ages, augmented by a stochastic particle set for matching cost estimation. Furthermore, a novel track-to-detection association strategy is proposed to reduce space complexity while preserving reliability and supporting group tracking. The pipeline leverages prior tracklets, particle states, current detections, and detection confidences, which executes track-detection matching through four stages: (1) clustering-based track–detection grouping, (2) valid masking and cost computation, (3) shared detection handling with local cost refinement, and (4) final association. A modified neighbor-based refinement further mitigates ambiguities in crowded scenes. A reference implementation is provided to validate the framework’s effectiveness.

The remainder of this paper is structured as follows: \Cref{sec:formatting} outlines the tracking pipeline, followed by the data association for temporal and scalable numbers of targets. \Cref{sec:evaluations} evaluates the proposed tracker on MOT17 and MOT20. Finally, the conclusions are presented in \Cref{sec:conclusions}.


\section{Methodology}
\label{sec:formatting}

To effectively address non-linear dynamics and non-Gaussian noise, multi-object tracking can be formulated as a particle filtering problem. Here, the system state is represented as $\{K_t, X_t\} = \{K_{i,t}, X_{i,t}\}_{i=1}^k$, with $K_t$ denotes a set of target identifiers of $k$ time-varying number of targets, and $X_t$ denotes their states. Then, the posterior density $P(K_t, X_t | Z_{0:t})$ is approximated using \textit{S} random samples. Unlike conventional methods, the proposed approach omits track creation and removal during sampling, employing instead a deterministic data association to ensure output consistency. This design resolves the challenges of internal model of jointly predicting target motion and preserving identity, thereby eliminating the need for full posterior sampling and focusing on optimizing particles for each target \cite{toan2025pami, toan2025icra}. In this paper, target particles are sampled via a MCMC approach, as $\{K_{t}^{S}, X_{t}^{S}\} \triangleq \sum_{s} P(K_t, X_t | K_{t-1}^{s},X_{t-1}^{s})$, and refined using a PSO algorithm \cite{kennedy1995particle, clerc2002PSO} with a specifically designed fitness function. The optimized particles are subsequently passed through the proposed track-detection matching to update target states.

The target state at time $t$ is presented by a bounding box $X_{i,t} = (u_{i,t}, v_{i,t}, w_{i,t}, h_{i,t})$, characterized by center $(u_{i,t}, v_{i,t})$ and size $(w_{i,t}, h_{i,t})$. Its corresponding velocity will be $V_{i,t} = (\dot{u}_{i,t}, \dot{v}_{i,t}, \dot{w}_{i,t}, \dot{h}_{i,t})$. The target velocities $V_t = \{V_{1,t}, ..., V_{k,t}\}$ are constrained by $V_t^{max} = \{V_{1,t}^{max}, ..., V_{k,t}^{max}\}$, satisfying $V_{i,t} \in (-V_{i,t}^{max}, V_{i,t}^{max})$, with $V_{i,t}^{max} = (\dot{u}_{i,t}^{max}, \dot{v}_{i,t}^{max}, \dot{w}_{i,t}^{max}, \dot{h}_{i,t}^{max})$. For the tracking inference in this paper, target tracklets are then formulated by their identifiers, states, velocities, penalties, and ages up to the current frame $I_t$, as $\{K_t, X_t, V_t, X_t^{pen}, X_t^{age}\} = \{K_{i,t}, X_{i,t}, V_{i,t}, X_{i,t}^{pen}, X_{i,t}^{age}\}_{i=1}^k$.

The following sections outline the tracking pipeline and the proposed track-detection matching for scalable numbers of targets. 

\subsection{Tracking pipeline}

\begin{figure}[t]
    \centering
    \includegraphics[width=0.9\columnwidth]{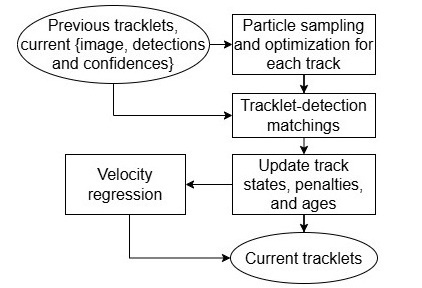}
    \caption{Overview of the tracking pipeline}
    \label{fig1}
\end{figure}

The tracking pipeline illustrated in \Cref{fig1} comprises three primary stages: particle sampling and optimization, track-detection association, and tracklet updates. Without an explicit motion model, particle sampling employs the random motion model in \Cref{equation1}: 

\begin{equation}
\label{equation1}
\begin{cases}
    V_{i,t} = V_{i,t-1} + \varepsilon_V.U_{V_{i,t}} \\
    X_{i,t} = X_{i,t-1} + \lambda_V.V_{i,t} + \lambda_X.\varepsilon_X.U_{X_{i,t}} \\
\end{cases}
\end{equation}

Parameters $\varepsilon_X, \varepsilon_V, \lambda_X$, and $\lambda_V$ regulate state and velocity explorations, with $\lambda_X + \lambda_V = 1$. Perturbations $U_{X_{i,t}} = (U_{i,t}^u, U_{i,t}^v, U_{i,t}^w, U_{i,t}^h)$ and $U_{V_{i,t}} = (U_{i,t}^{\dot{u}}, U_{i,t}^{\dot{v}}, U_{i,t}^{\dot{w}},  U_{i,t}^{\dot{h}})$ bounded by $(-U_{X_{i,t}}^{max}, U_{X_{i,t}}^{max})$ and $(-U_{V_{i,t}}^{max}, U_{V_{i,t}}^{max})$ which are determined by its bounding box size in the previous state $\{K_{i,t-1}, X_{i,t-1}, V_{i,t-1}\}$. To guide particles sampled from \Cref{equation1} toward optimal positions, PSO algorithm is applied with a fitness function combining history fitness $f_{PSO}^h$, exploration fitness $f_{PSO}^p$, and social fitness $f_{PSO}^i$.      

\begin{equation}
\label{equation2}
    f_{PSO} = \sigma_h.f_{PSO}^h + \sigma_p.f_{PSO}^p + \sigma_i.f_{PSO}^i 
\end{equation}

Here, weights $\sigma_h, \sigma_p$, and $\sigma_i$ are positive and $\sigma_h + \sigma_p + \sigma_i = 1$. Both $f_{PSO}^h$ (comparing the particle to the previous optimal state) and $f_{PSO}^p$ (comparing it to its prior PSO update) are measured by $f(\bullet, \bullet)$, as:

\begin{equation}
\label{equation3}
    f(\bullet, \bullet) = \lambda_s.f_s + \lambda_m.f_m
\end{equation}

Here, $f_s \in [0, 1]$ denotes the cosine similarity between non-negative HoG features of two input bounding boxes, while motion fitness $f_m$ measures their distance, normalized to $[0, 1]$ by a target-adaptive maximum $d_{o,m}$. Weights $ \lambda_s$ and $\lambda_m$ are positive and $\lambda_s + \lambda_m = 1$.

Target interactions impact performance of multi-object tracking, especially under occlusions. The neighbour set $\{K_{i,t}, X_{i,t}\}^{nei} = \{K_{j,t}^n, X_{j,t}^n\}_{n=1}^N$ of target $\{K_{i,t}, X_{i,t}\}$ is obtained via a nearest-neighbour search $\{K_{i,t}, X_{i,t}\}^{nei} = \Psi\langle (K_{i,t}, X_{i,t}), \varepsilon_{nei} \rangle$, with adaptive threshold $\varepsilon_{nei}$ based on the size of the bounding box of $X_{i,t}$, which does not directly participate in data association. Instead, social fitness is employed within PSO to guides particles to diverge from nearby states, reducing ID switches, as computed as in \Cref{equation4}, with $|x|$ denotes the magnitude of the vector $x$. Here, $V_s^{max} = V_{i,t}^{max} + U_{V_{i,t}}^{max}$, with $\xi_p + \xi_V = 1$. And, if $N=0, f_{PSO}^{i,s} = 1$.   

\begin{equation}
\label{equation4}
\begin{split}
     &f_{PSO}^{i,s} = \frac{\xi_p}{N}.\sum_{j=1}^N \frac{min(|X_{i,t}^s - X_{j,t}|, 2\varepsilon_{nei})}{2\varepsilon_{nei}} \\ %
     &+ \frac{\xi_V}{N}.\sum_{j=1}^N \frac{min(|V_{i,t}^s - V_{j,t}|, V_s^{max})}{V_s^{max}}
\end{split}
\end{equation}

Optimized particles are then used for track-detection data association via the Hungarian algorithm \cite{kuhn1955hungarian} with the target-oriented cost matrix $C_{mat} \in \mathbb{R}^{T \times D}$, where $T$ and $D$ are the numbers of targets and detections, ($T = k$). The matching cost $C_{i,j}$ between $X_{i,t}$ and detection $det_j$, is computed by using $S$ particles of $X_{i,t}$, track penalty $X_{i, t-1}^{pen} \in [0, 1]$, and detection confidence $det_j^{conf} \in [0, 1]$:

\begin{equation}
    \label{equation5}
    \begin{split}
        C_{i,j} = \lambda_p.\frac{1}{S}. \sum_{s=1}^S C_m^{X_{i,t}^s,det_j}  + \lambda_d.(1-det_j^{conf}) \\ %
        + \lambda_h.X_{i,t-1}^{pen}
\end{split}
\end{equation}

Here, $\lambda_h, \lambda_d$ and $\lambda_p$ are positive and $\lambda_h + \lambda_d + \lambda_p = 1$. The motion consistent cost $C_m^{X_{i,t}^s,det_j} \in [0, 1]$ is re-defined as:       

\begin{equation}
\label{equation6}
    C_m^{X_{i,t}^s,det_j} = \sigma_u \cdot C_{IoU}^{X_{i,t}^s,det_j} + \sigma_d \cdot C_d^{X_{i,t}^s,det_j}
\end{equation}

The IoU cost is defined as $C_{IoU}^{X_{i,t}^s, det_j} = 1 - IoU_{det_j}^{X_{i,t}^s}$. The distance cost $C_d^{X_{i,t}^s,det_j}$ quantifies the track-detection bounding box separation, normalized to $[0, 1]$ using a maximum distance derived from their current bounding box sizes, with $det_j = (u_{det}, v_{det}, w_{det}, h_{det})$. It is noted that $IoU_{det_j}^{X_{i,t}^s}$ is computed as a combination of the two-dimensional spatial intersection and height-based intersection, and $\sigma_u + \sigma_d = 1$.  

By using $C_{mat}$, a set of M pairs $\{K_{i,t}^m, X_{i,t}^m,det_{j,t}^m\}_{m=1}^M$ is identified. Strong tracks are then defined as $(K_{i,t}^m, X_{i,t}^m)$. Weak tracks are residual set $\{K_t^w, X_t^w\} = \{K_t, X_t\} - \{K_{i,t}^m, X_{i,t}^m\}$. Unmatched detections $\{det_t^u, det_t^{u,conf}\} = \{D_t, D_t^{conf}\} - \{det_t^m, det_t^{m, conf}\}$ are used to initialize new tracks $\{K_t^n, X_t^n\}$. The state of system now can be detailed as: $\{K_t, X_t\} = \{(K_t^m, K_t^w, K_t^n), (X_t^m, X_t^w, X_t^n)\}$.

Track states are updated according to their types, with $X_t^m$ and $X_t^n$ adopting their associated detection states, and penalties $X_t^{m,pen}, X_t^{n,pen}$ and ages $X_t^{m,age}, X_t^{n,age}$ reset to zero. Weak tracks $(K_t^w, X_t^w)$, lacking matched detections, are updated using their PSO global bests and neighbours $\{K_{i,g}, X_{i,g}, f_{i,g}, X_i^{nei}\}_{i=1}^k$, with neighbors’ states taken from matched detections if strong, or from previous optimal states otherwise. States of weak tracks are then updated as: 

\begin{equation}
\label{equation7}
\begin{cases}
    \{K_t^w,X_t^w\} \xleftarrow{} \{K_{g}^w,X_{g}^w,f_{g}^{w}, V_t^w, X_w^{nei}\} \\
    X_t^{w,pen} = X_{t-1}^{w,pen} + \zeta.\Delta_t \\
    X_t^{w,age} = X_{t-1}^{w,age} + \zeta.\Delta_t.\partial_{max}\\
    \Delta_t = (1 - e^{\frac{-l^2}{2\sigma^2}}).(1-f_{g}^{w} + \Delta_e)\\
\end{cases}
\end{equation}

Here, $\partial_{max}$ denotes the maximum age, and $\zeta$ represents recovery trust, defined as $\zeta = sign(\rho_{re} - f_{g}^{w} + \Delta_e)$ if a weak track has strong neighbours, and $\zeta = 1$ otherwise. $\Delta_e$ is an optional entrance penalty defined inside custom entrance areas, $l$ counts consecutive unmatched frames, $\sigma = \eta.\partial_{max}$ with $\eta > 1$, and $\rho_{re} \in [0, 1]$ is the recovery threshold. Track penalties and ages satisfy $1 \ge X_t^{w,pen} \ge 0$, and $ \partial_{max} \ge X_t^{w,age} \ge 0$. It is noted that updating bounding box size like its center location during occlusions causes scale drift; thus, position and size updates should be decoupled. For deeper insight into the tracking pipeline, it is encouraged to consult \cite{toan2025pami, toan2025icra}. 

Beyond PSO social fitness, neighbour $(K_{j,t}, X_{j,t})$ influences weak track $(K_t^w, X_t^w)$ based on its reliability: strong-track neighbour is reliable to contribute to state update, whereas weak-track neighbour is excluded. Let $\{X_t^b, V_t^b \}$ denote the set of reliable neighbours and $\{\overline{X}_t^b, \overline{V}_t^b \}$ denote their median. Unlike \cite{toan2025pami, toan2025icra}, the state of weak track in this paper is updated as: 

\begin{equation}
\label{equation8}
\begin{cases}
    X_t^w = \overline{X}_t^b - \overline{X}_{t-1}^b + X_{t-1}^w, & \text{if } \delta \ge \delta_d   \\
    X_t^w = (1-\sigma_g).\hat{X}_{t}^w+ \sigma_g.X_g^w, & \text{otherwise} \\
\end{cases}
\end{equation}






Here, $\delta$ denotes the cosine similarity between weak-track and neighbour velocities, with $\delta_d$ is the similarity threshold. $X_g^w$ and $\sigma_g$ represent the weak track’s post-PSO global best and its associated trust factor, while $\hat{X}_{t}^w$ is updated by using its previous optimal state and velocity. The size of $\hat{X}_{t}^w$ remains similar as of $X_{t-1}^w$, while the bounding box center's position ${}^{[u,v]}\hat{X}_t^w$ is updated as:  

\begin{equation}
    \label{equation9}
    {}^{[u,v]}\hat{X}_t^w = {}^{[u,v]}X_{t-1}^w + {}^{[u,v]}V_t^w  
\end{equation}




It is noted that the weak track state is updated only if $|{}^{[u,v]}V_t^w| \ge \tau_V$ to suppress noise, with size adjustments made minimally or upon a detection match. After updates, penalties $\{X_t^{pen}\} = \{X_t^{m, pen}, X_t^{w, pen}, X_t^{n, pen}\}$ and ages $\{X_t^{age}\} = \{X_t^{m, age}, X_t^{w, age}, X_t^{n, age}\}$ of all targets are recorded, and expired tracks are removed from the tracklets as follows:   

\begin{equation}
\label{equation10}
\begin{split}
    & \{K_t,X_t,V_t,X_t^{pen},X_t^{age}\} = \\ %
    & \{K_{i,t},X_{i,t},V_{i,t},X_{i,t}^{pen},X_{i,t}^{age} | X_{i,t}^{age} < \partial_{max}\}_{i=1}^k
\end{split}
\end{equation}

Finally, for each current target $X_{i, t}$, velocity regression over its past $H$ states with a frame window $F$ $(0 < F \le H)$ estimates trend-based velocity for subsequent particle generation and state updates.

The presented tracking pipeline enables robust performance under nonlinear target dynamics. Nonetheless, scalability remains a critical issue due to the reliance on a full cost matrix $C^{TxD}$ for global data association, with $T$ and $D$ are total numbers of current targets and detections. It incurs substantial computational overhead and hinders real-time applicability when the number of targets significantly increases. While some prior works \cite{tingpeng2016improved, ayokor2007dynamics}, not MOT-specific, have introduced modified Hungarian algorithms and incremental cost updates, they do not adequately address the raised limitation. In this paper, a novel data association framework is introduced that preserves global track-detection matchings while improving scalability. Distinct from multi-stage matching strategies \cite{csaba2010multi, zhen2012improving, wojke2017simple, yifu2022bytetrack}, the proposed method systematically constructs a cost matrix that partitions the global matching problem into smaller, localized associations within a single-stage process. Global optimality is subsequently maintained through shared detection processing. The detailed formulation of the proposed track-detection matching approach is provided in the following section.    

\subsection{Proposed track-detection matching}

\begin{figure}[t]
    \centering
    \includegraphics[width=0.9\columnwidth]{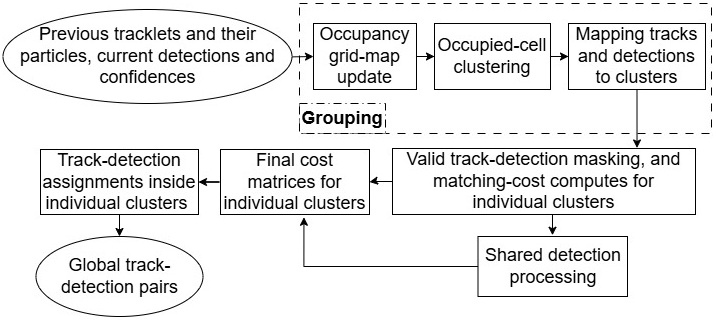}
    \caption{Pipeline for track-detection matching}
    \label{fig2}
\end{figure}

To address the spatial complexity of the matching cost matrix while supporting group tracking, an association framework is introduced. The track-detection pipeline, illustrated in \Cref{fig2}, utilizes previous target tracklets and particle states, along with current detections and associated confidences. Following input acquisition, the pipeline comprises four stages: track-detection grouping, valid track-detection masking and cost computation, shared detection processing with local cost matrix refinement, and final track-detection matchings. 

\begin{figure*}[!t]
    \centering
    \includegraphics[width=\textwidth]{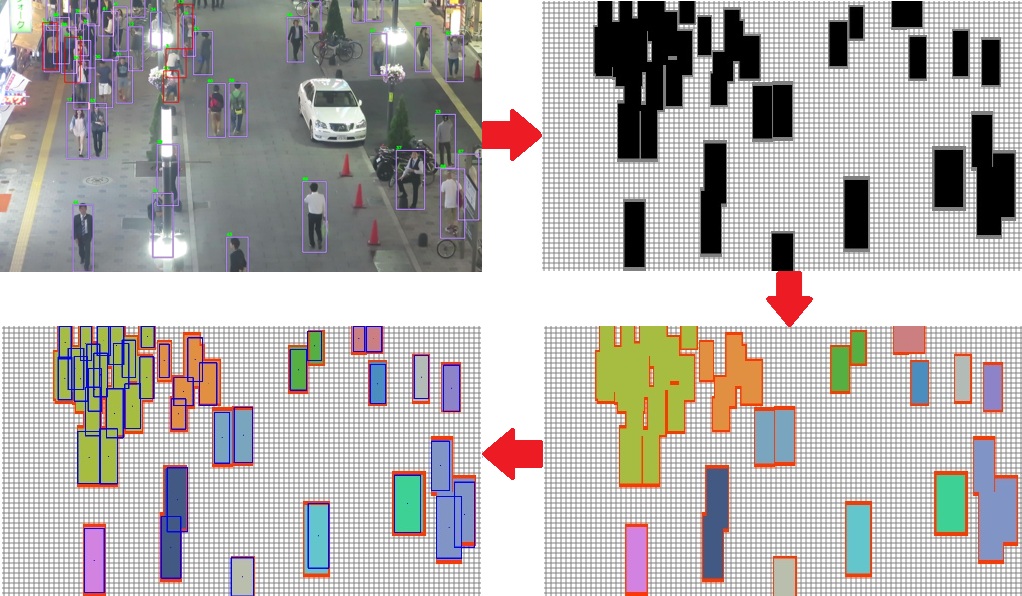}
    \caption{Track and detection grouping sampled from the MOT17-04 sequence \cite{dendorfer2021motchallenge}, containing 45 targets in the current frame. Top-left: current tracking scene; top-right: updated 2-D occupancy grid map; bottom-right: clustering results, with different colors representing distinct clusters; bottom-left: mapping of tracks and detections to their respective clusters. The blue bounding boxes indicate detections.}
    \label{fig3}
\end{figure*}

The first stage clusters tracks and detections to partition the global cost matrix into smaller, independent submatrices for efficient matching, as illustrated in \Cref{fig3}. A two-dimensional (2-D) occupancy grid map $G$ is defined over the image $I_t$ with width $I_w$ and height $I_h$, as $G=\{g_{y,x} | 0 \le y \le \frac{I_h}{g_y}, 0 \le x \le \frac{I_w}{g_x} \}$, where $g_x=n.I_p$ and $g_y=m.I_p$ denote the grid cell dimensions along the width and height, and $I_p$ is the image pixel unit. In this paper, the grid-map was configured with $g_x = 5$ pixels and $g_y = 5$ pixels. Accordingly, the grid-map dimensions are $(G_w, G_h) = (\frac{I_w}{g_x}, \frac{I_h}{g_y})$. The occupancy of each cell is updated based on the previous optimal target states, as:      

\begin{equation}
\label{equation11}
g_{y,x} =
\begin{cases}
    1, &\text{if } y \in B_y \And x \in B_x   \\
    0, &\text{otherwise}\\
\end{cases}
\end{equation}

For each target state $X_i = (u_i, v_i, w_i, h_i)$ with $ \forall X_i \in X_{t-1}$, $B_x = [\frac{u_i - 0.5w_i}{g_x} - \varepsilon.\frac{w_i}{g_x}, \frac{u_i + 0.5w_i}{g_x} + \varepsilon.\frac{w_i}{g_x}]$ and $B_y = [\frac{v_i - 0.5h_i}{g_y} - \varepsilon.\frac{h_i}{g_y}, \frac{v_i + 0.5h_i}{g_y} + \varepsilon.\frac{h_i}{g_y}]$, where $\varepsilon$ is the inflation ratio. Since all bounding boxes lie within the image boundaries, $\{B_y,B_x\} \subseteq \{G_y, G_x\}$. The set of occupied cells is $G_{occ} = \{g_{y,x} | g_{y,x} = 1\}$, and the set of free cells is the complement $G_{free} = G - G_{occ}$. The grid-map update process is illustrated in the top-right of \Cref{fig3}. Subsequently, clustering is performed on the 2-D occupancy grid map $G$ using connected-component labeling. Cells in $G_{free}$ are assigned to the background cluster and excluded from subsequent processing. Valid clusters are extracted from $G_{occ}$ by evaluating 8-connected neighborhoods, where adjacent occupied cells are grouped into the same cluster. The clustering process is illustrated in the bottom-right of \Cref{fig3}, with distinct clusters visualized in different colors. Each cluster comprises occupied cells belonging to the same connected region. After the cluster set $L=\{L_{occ}\}$ is obtained, tracks and detections are mapped to their respective clusters. Since the occupancy grid is updated using the bounding boxes of track states, a track $X_i$ is assigned to cluster $L_{occ}$ if $(\frac{v_i}{g_y}, \frac{u_i}{g_x}) \in L_{occ}$, ensuring a unique cluster association per track. Detections, however, are subject to noise and temporal discrepancies between frames. Thus, a detection $det_j = (u_j, v_j, w_j, h_j)$ is assigned to cluster $L_{occ}$ if $(\frac{v_j}{g_y}, \frac{u_j}{g_x}) \in L_{occ}^{exp}$, where $L_{occ}^{exp}$ denotes an expanded region of $L_{occ}$ by half the detection bounding box size. This expansion may result in detections belonging to multiple clusters, which is resolved in the subsequent stage.

In the second stage, after assigning tracks and detections to clusters, a valid track-detection pair mask is constructed for each cluster to eliminate redundant cost computes. Since matches are implausible when tracks and detections are spatially distant, and computing full matching costs is more expensive than simple distance calculation, the mask effectively reduces computational complexity - particularly with large target sets. For cluster $L_{occ}$, the matching cost matrix is defined as $C^L \in R^{T^L x D^L}$, where $T^L$ and $D^L$ denote the numbers of tracks and detections within the cluster. The corresponding valid-pair mask $M^L \in R^{T^L x D^L}$ is updated as follows:     

\begin{equation}
\label{equation12}
M_{i,j}^L =
\begin{cases}
    1, &\text{if } |X_i^L - det_j^L| \le \lambda_D.\sqrt{w_j^2 + h_j^2}   \\
    0, &\text{otherwise}\\
\end{cases}
\end{equation}

\begin{figure}[t]
    \centering
    \includegraphics[width=0.43\columnwidth]{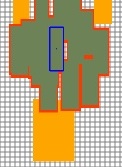}
    \caption{Example of a valid-pair mask within an individual cluster for a specific detection. The blue bounding box indicates the detection.}
    \label{fig4}
\end{figure}

Here, $X_i^L$ and $det_j^L$ denote a track and a detection assigned to cluster $L_{occ}$, with $w_j$ and $h_j$ representing the width and height of $det_j^L$. The factor $\lambda_D > 1$ controls the valid spatial range around a detection. \Cref{fig4} illustrates the valid track–detection mask for a cluster: a track is valid for matching with the detection if its bounding box region is entirely green, and invalid if any portion is orange. For invalid pairs ($M_{i,j}^L = 0$), the matching cost $C_{i,j}^L$ is assigned a large value; otherwise, the cost is computed using \Cref{equation5} and \Cref{equation6}.

\begin{figure}[t]
    \centering
    \includegraphics[width=0.95\columnwidth]{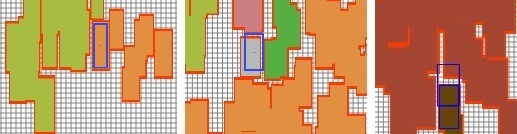}
    \caption{Example scenarios where a detection is assigned to multiple clusters. The blue bounding box highlights the detection, while different colors represent distinct clusters.}
    \label{fig5}
\end{figure}

\begin{figure}[t]
    \centering
    \includegraphics[width=0.95\columnwidth]{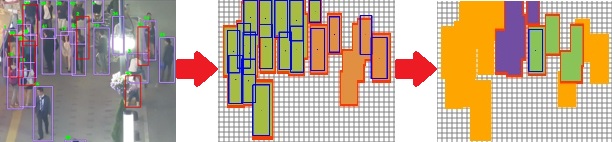}
    \caption{A shared detection - shown as a blue bounding box (last part of figure)- is initially assigned to two different clusters before being correctly finalized in the right-side cluster.}
    \label{fig6}
\end{figure}

One track is always assigned to a unique cluster, whereas a detection may belong to multiple clusters, as shown in \Cref{fig5}.  For global matching, a detection cannot be assigned to multiple tracks. To enforce this, shared-detections undergo post-processing: each is reallocated to the optimal cluster to ensure a one-to-one track-detection correspondence. Specifically, all valid tracks associated with a shared-detection are compared, and the detection is allocated to the best-matching cluster. Given that valid masks constrain the comparisons, the number of candidate pairs remains small.

For a shared detection $det^s$, let $\{L_l^s\}_{l=1}^E$ denote its associated list of $E$ clusters. For each cluster $L_l^s$, the best track and corresponding cost for $det^s$ are determined. Let  $\{D_t^L\}$ represent the set of all detections in $L_l^s$, so that $det^s \in \{D_t^L\}$. A track $X_l^s \in \{X_l^L\}$ is defined as the best match of $det^s$ in $L_l^s$ if it satisfies $\forall det_j^L \in \{D_t^L\}, C_{X_l^s}^{det_j^L} > C_{X_l^s}^{det^s} \land \forall X_{l,i}^L \in \{X_l^L\}, C_{X_{l,i}^L}^{det^s} > C_{X_l^s}^{det^s}$, where $\{X_l^L\}$ is the set of all valid pairwise tracks associated with $L_l^s$, and $C_X^{det}$ denotes the matching cost between track $X$ and detection $det$. Then, the representative cost of the shared detection $det^s$ to cluster $L_l^s$ is defined as $C_{L_l^s}^{det^s} = C_{X_l^s}^{det^s}$, and it is used to compare $det^s$ across clusters. If $\neg \exists X_l^s \in \{X_l^L\}$ satisfies the previous condition, $C_{L_l^s}^{det^s}$ is assigned a very big value. After the representative costs of $det^s$ across its associated clusters are determined, it is assigned to the cluster $L_b^s$ that minimizes the cost, as $C_{L_b^s}^{det^s} < C_{L_q^s}^{det^s}, \forall L_q^s \in \{L_l^s\} \land b \neq q$. Finally, the matching costs of $det^s$ for tracks in all other clusters are then set to a large value. An illustration of the shared-detection process is provided in \Cref{fig6}.

After shared detections are assigned to their optimal clusters, final cost matrices for track-detection matching within each cluster are established. Local track-detection assignments are then performed per cluster. Although track-detection assignments are performed locally, the overal matching remains global through the pipeline illustrated in \Cref{fig2}. This approach also enhances conditions for group tracking.

The primary differences in complexity between the proposed track-detection matching method and the approaches in \cite{toan2025pami, toan2025icra} arise from cost matrix computation and assignment. For a global cost matrix of $T$ tracks and $D$ detections, as in \cite{toan2025pami, toan2025icra}, the space complexity is $O(T \cdot D)$ for cost matrix and $O(T \cdot D + T + D)$ for the assignments. In contrast, for the proposed method, which partitions tracks and detections into $L$ clusters, the space complexity of the cost matrices and assignments is $\sum_{l=1}^L{O(T_l \cdot D_l)}$ and $\sum_{l=1}^LO(T_l \cdot D_l+T_l+D_l)$, respectively, where  $T_l \ge 0$ and $D_l \ge 0$ denote the number of tracks and detections in the $l^{th}$ cluster. Since $T_l$ and $D_l$ are non-negative and $T = \sum_{l=1}^L T_l$ and $D =  \sum_{l=1}^L D_l$, it is straightforward to verify that $T \cdot D \ge \sum_{l=1}^LT_l \cdot D_l$. Therefore, $O(T \cdot D) \ge \sum_{l=1}^L O(T_l \cdot D_l)$ and $O(T \cdot D + T + D) \ge \sum_{l=1}^L O(T_l \cdot D_l + T_l +D_l)$. The time complexity of constructing the global cost matrix is $O(T \cdot D \cdot t_c)$, where $t_c$ denotes the computation time per cost, as defined in \Cref{equation5}, while the worst case time complexity of the assignment step is $O(max(T,D)^3)$. Analogously to the space complexity analysis, the time complexity of the global cost matrix exceeds that of the proposed method. Furthermore, applying valid track-detection masks reduces the complexity of the cost matrix. A single cost computation in \Cref{equation5} involves $S$ IoU evaluations and $S$ box distance computations, in addition to other operations, while the mask requires only a single box distance per entry. Consequently, masking reduces both computational time and resource usage.

Increasing the number of clusters reduces the computational cost of cost matrix construction and assignment by operating on smaller subsets of tracks and detections. However, this may increase the complexity of shared detection processing. The computational complexity of both the grouping stage and shared detection processing is further influenced by the density of the tracking scenario. Further analysis follows in \Cref{sec:evaluations}.  

Although the shared-detection processing generally ensures global assignments, a near-optimal solution can be obtained for stricter computation by selecting the $top_n$ tracks from each cluster rather than all valid tracks, with $top_n = 1$ using direct mutual lowest-cost. In practice, this approach is more scalable while remaining near-optimal since each detection is assigned to the cluster with the lowest achievable cost and the global solution would not assign a detection to a higher-cost cluster.

\section{Evaluations}
\label{sec:evaluations}

\begin{figure}[t]
    \centering
    \includegraphics[width=0.95\columnwidth]{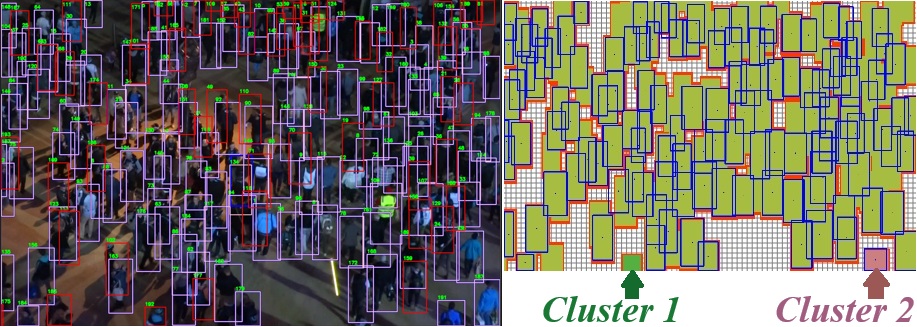}
    \caption{Crowded scene from MOT20-05 sequence \cite{pattrick2020mot20} showing 188 targets in the current frame. Three clusters are identified: clusters 1 and 2 each contain a single target, while the last cluster contains the remaining 186 targets.}
    \label{fig7}
\end{figure}

The proposed method was evaluated on the MOT17 \cite{dendorfer2021motchallenge} and MOT20 \cite{pattrick2020mot20} datasets. Specifically, the MOT17-04 sequence (including 1050 frames at a resolution 1920×1080 and was recorded at 30 frames per second (FPS)) was used to assess performance in sparse scenes, while the MOT20-05 (including 3315 frames at a resolution 1654x1080 and was recorded at 25 FPS) was employed for crowded scenarios. Illustrations of these sequences are shown in the top-left of \Cref{fig3} and in the left of \Cref{fig7}, respectively. 



Furthermore, comparative evaluations were conducted against previous trackers, including SORT \cite{bewley2016simple}, DeepSORT \cite{wojke2017simple}, ByteTrack \cite{yifu2022bytetrack}, BoT-SORT \cite{nir2022botsort}, OC-SORT \cite{jinkun2023OCSORT}, SMILEtrack \cite{yu2024smiletrack}, ConfTrack \cite{hyeonchul2024ConfTrack}, BoostTrack \cite{vukas2024inboostrack}, GenTrack Super \cite{toan2025pami}, and GenTrack2 Super \cite{toan2025icra}, using identical detections and ground truths. The GenTrack framework introduced in \cite{toan2025pami} features three variants, among which the full version, GenTrack Super, is used for comparison. The method in \cite{toan2025icra}, being an extension of GenTrack Super, is referred to as GenTrack2 Super. The proposed approach represents a further extension, will be termed as GenTrack3 Super. Additionally, a variant, GenTrack2 Super+, is evaluated; it employs the complete GenTrack2 Super pipeline but replaces its track-detection matching component with the proposed matching method in this paper.  

All experiments were performed on a PC equipped with an AMD Ryzen 7 PRO 8840U processor, 16GB RAM, without dedicated GPU (Graphics Processing Unit). A reference implementation of the proposed method, along with re-evaluable implementations of the compared trackers, is provided for re-evaluation. Tracking performance was assessed using established metrics \cite{vasant2006performance, luiten2021hota}, including ATA (Average Tracking Accuracy), IDF1, HOTA (Higher Order Tracking Accuracy), and MOTA (Multiple Object Tracking Accuracy). Here, their values range from 0 to 100, with higher values indicate better results.


\begin{table}
  \caption{Tracking Performance Evaluated on the MOT17-04 \cite{dendorfer2021motchallenge}}
  \label{tab:table1}
  \centering
  \begin{tabular}{@{}lc c c c @{}c} 
    \toprule
   
    {\textbf{Tracker}} & {\textbf{ATA}} & {\textbf{IDF1}} & {\textbf{HOTA}} & {\textbf{MOTA}} \\ 

    \midrule
    {SORT \cite{bewley2016simple}} & {56.17} & {82.60} & {61.96} & {69.86} \\ 

    {DeepSORT \cite{wojke2017simple}} & {51.00} & {89.44} & {55.51} & {75.00} \\ 

    {ByteTrack \cite{yifu2022bytetrack}} & {49.73} & {83.28} & {62.94} & {73.84} \\ 

    {BoT-SORT \cite{nir2022botsort}} & {40.34} & {82.79} & {61.74} & {70.88} \\ 

    {OC-SORT \cite{jinkun2023OCSORT}} & {49.59} & {81.41} & {59.59} & {69.11} \\ 

    {SMILEtrack \cite{yu2024smiletrack}} & {44.10} & {83.97} & {62.17} & {71.48} \\ 

    {ConfTrack \cite{hyeonchul2024ConfTrack}} & {51.75} & {90.18} & {68.10} & {82.16} \\ 

    {BoostTrack \cite{vukas2024inboostrack}} & {49.88} & {81.50} & {61.34} & {69.40} \\ 

    {GenTrack Super \cite{toan2025pami}} & {62.95} & {93.59} & {68.60} & {85.11} \\ 

    {GenTrack2 Super \cite{toan2025icra}} & {66.90} & {\textbf{93.88}} & {\textbf{70.04}} & {85.72} \\ 

    {\textbf{GenTrack2 Super+}} & {\textbf{68.94}} & {93.40} & {66.64} & {\textbf{87.31}} \\ 
    
    {\textbf{GenTrack3 Super}} & {63.62} & {93.16} & {67.61} & {84.86} \\ 
    
    \bottomrule
  \end{tabular}
\end{table}

\begin{table}
  \caption{Tracking Performance Evaluated on the MOT20-05 \cite{pattrick2020mot20}}
  \label{tab:table2}
  \centering
  \begin{tabular}{@{}lc c c c @{}c} 
    \toprule
   
    {\textbf{Tracker}} & {\textbf{ATA}} & {\textbf{IDF1}} & {\textbf{HOTA}} & {\textbf{MOTA}} \\ 

    \midrule
    {SORT \cite{bewley2016simple}} & {29.63} & {67.01} & {38.19} & {49.93} \\ 

    {DeepSORT \cite{wojke2017simple}} & {22.47} & {84.47} & {26.11} & {73.68} \\

    {ByteTrack \cite{yifu2022bytetrack}}& {20.93} & {63.89} & {37.03} & {55.25} \\

    {BoT-SORT \cite{nir2022botsort}} & {14.14} & {63.61} & {37.98} & {55.51} \\ 

    {OC-SORT \cite{jinkun2023OCSORT}} & {22.53} & {65.56} & {40.94} & {55.24} \\ 

    {SMILEtrack \cite{yu2024smiletrack}} & {20.96} & {70.16} & {40.07} & {55.62} \\ 

    {ConfTrack \cite{hyeonchul2024ConfTrack}} & {28.23} & {81.20} & {\textbf{45.59}} & {78.39} \\ 

    {BoostTrack \cite{vukas2024inboostrack}} & {23.98} & {62.69} & {38.08} & {50.00} \\ 

    {GenTrack Super \cite{toan2025pami}} & {28.02} & {88.00} & {34.27} & {75.14} \\ 

    {GenTrack2 Super \cite{toan2025icra}} & {\textbf{35.84}} & {\textbf{91.88}} & {39.96} & {77.45} \\ 

    {\textbf{GenTrack2 Super+}} & {35.16} & {91.10} & {39.82} & {76.89} \\ 

    {\textbf{GenTrack3 Super}} & {34.00} & {88.23} & {44.23} & {\textbf{78.72}} \\ 
    
    \bottomrule
  \end{tabular}
\end{table}

The results, summarized in \Cref{tab:table1} and \Cref{tab:table2}, with best metrics in bold, demonstrate distinct performance pattern across trackers. In \Cref{tab:table1}, GenTrack2 Super achieves the highest IDF1 and HOTA scores, while GenTrack2 Super+ attains superior ATA and MOTA values. However, \Cref{tab:table2} reveals trade-offs between identity consistency and localization: GenTrack2 Super excels in ATA and IDF1, whereas ConfTrack achieves the highest HOTA, and GenTrack3 Super achieves the highest MOTA score. These findings indicate that the weak track refinement mechanism in GenTrack2 Super effectively manages occlusions and ensures trajectory smoothness in sparse scenes but can cause state shifts in dense environments, such as the MOT20-05, where weak tracks are surrounded by numerous others. The performance of GenTrack2 Super+ further demonstrates that the proposed track-detection matching strategy effectively reduces computational complexity while maintaining comparable global assignment accuracy. Overall, the complete framework in this paper, GenTrack3 Super, achieves a competitive performance across both sparse and crowded scenes. 

\begin{figure}[t]
    \centering
    \includegraphics[width=\columnwidth]{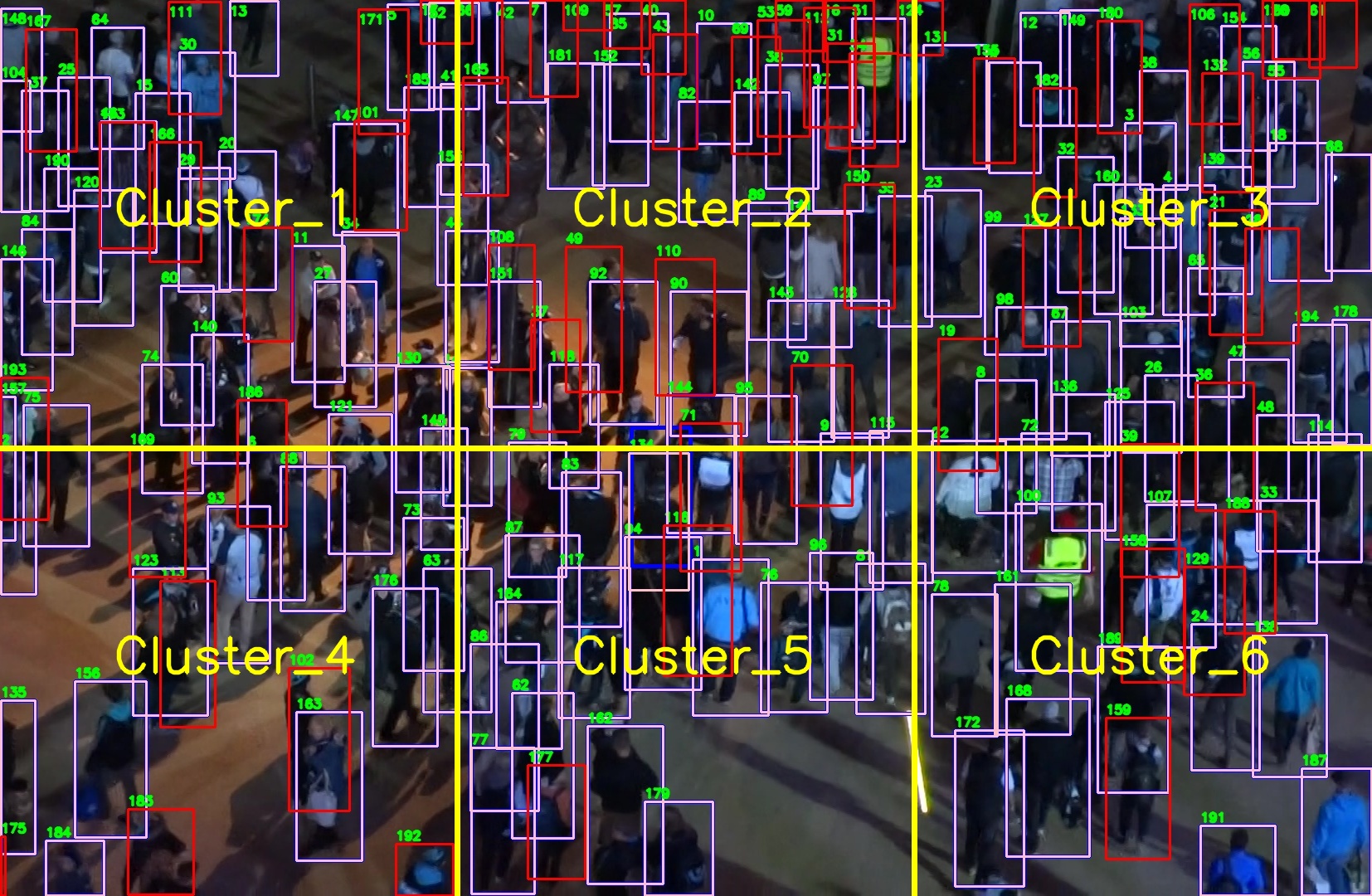}
    \caption{Image-region clustering: the image is partitioned into six smaller regions, each treated as an individual cluster.}
   \label{fig8}
\end{figure}

Observations from implementation identify an edge case when a large number of targets densely occupy the observation region, as shown in \Cref{fig7}. In this case, target grouping using occupancy grid maps and clustering does not help to reduce the computation. Specifically, the right side of \Cref{fig7} shows three clusters: clusters 1 and 2 each contain a single target, whereas the last cluster encompasses 186 targets. In this scenario, track grouping is achieved by partitioning the image into smaller regions, with each region treated as a distinct cluster. The subsequent track-detection matching pipeline remains identical to the original approach, except that the occupancy grid map used for clustering is replaced by a simple spatial division. An example is shown in \Cref{fig8}, where the image is divided into six regions, each representing a separate cluster. 

However, regional clustering results in a higher number of shared detections, increasing the complexity of shared-detection processing and potentially reducing accuracy. These observations indicate that the proposed track-to-detection method is suited for sparse environments, such as robot navigation and human–robot interaction, where group tracking is essential for socially aware navigation. In contrast, the global masking cost matrix provides a better balance between performance in sparse and dense scenarios, offering greater overall robustness. A comprehensive analysis across GenTrack versions will be reported in future work.

\section{Conclusions}
\label{sec:conclusions}

This paper introduces an online visual MOT framework that integrates deterministic and stochastic components for robust and scalable tracking under uncertainty. Each target is modeled using a stochastic particle set for optimal state estimation, while a novel track-to-detection association preserves temporal identity consistency. The tracking inference employs tracklets encompassing identifiers, states, velocities, penalties, and ages to support a unified tracking pipeline. Insights from implementation indicate that the proposed track-to-detection method is suited for sparse environments, such as robot navigation and human–robot interaction, where group tracking is essential for socially aware navigation.    



\section*{Acknowledgment}

This work was a part of VelKoTek project, funded by The Ministry of Food, Agriculture and Fisheries of Denmark via the Green Development and Demonstration Program (GUDP).


\begin{thebibliography}{99}
\bibitem{csaba2010multi}
Csaba Beleznai and David Schreiber, "Multiple object tracking by hierarchical association of spatio-temporal data," in 2010 IEEE International Conference on Image Processing, 2010, pp. 41-44.

\bibitem{zhen2012improving}
Zhen Qin and Christian R. Shelton, "Improving Multi-target Tracking via Social Grouping," in 2012 IEEE Conference on Computer Vision and Pattern Recognition, 2012, pp. 1972-1978.

\bibitem{bewley2016simple}
Bewley Alex and Ge Zongyuan and Ott Lionel and Ramos Fabio and Upcroft Ben, "Simple online and realtime tracking," in 2016 IEEE International Conference on Image Processing (ICIP), 2016, pp. 3464-3468.

\bibitem{wojke2017simple}
Nicolai Wojke and Alex Bewley and Dietrich Paulus, "Simple online and realtime tracking with a deep association metric," in 2017 IEEE International Conference on Image Processing (ICIP), 2017, pp. 3645-3649.

\bibitem{yifu2022bytetrack}
Yifu Zhang and Peize Sun and Yi Jiang and Dongdong Yu and Fucheng Weng and Zehuan Yuan and Ping Luo and Wenyu Liu and Xinggang Wang, "ByteTrack: Multi-object tracking by associating every detection box," in Computer Vision – ECCV 2022. Lecture Notes in Computer Science, vol. 13682, 2022, Springer Cham.

\bibitem{nir2022botsort}
Nir Aharon and Roy Orfaig and Ben-Zion Bobrovsky, "BoT-SORT: Robust associations multi-pedestrian tracking," arXiv, 2022. Available at: \href{https://doi.org/10.48550/arXiv.2206.1465}{\url{https://doi.org/10.48550/arXiv.2206.14651}}

\bibitem{jinkun2023OCSORT}
Jinkun Cao and Jiangmiao Pang and Xinshuo Weng and Rawal Khirodkar and Kris Kitani, "Observation-Centric SORT: Rethinking SORT for robust multi-object tracking," in 2023 IEEE/CVF Conference on Computer Vision and Pattern Recognition (CVPR), 2023, pp. 9686-9696.

\bibitem{yu2024smiletrack}
Yu-Hsiang Wang and Jun-Wei Hsieh and Ping-Yang Chen and Ming-Ching Chang and Hung-Hin So and Xin Li, "SMILEtrack: Similarity learning for occlusion-aware multiple object tracking," in Proceedings of the Thirty-Eighth AAAI Conference on Artificial Intelligence, vol. 38, no. 6, pp. 5740-5748, 2024.

\bibitem{hyeonchul2024ConfTrack}
Hyeonchul Jang and Seokjun Kang and Takgen Kim and HyeongKi Kim, "ConfTrack: Kalman filter-based multi-person tracking by utilizing confidence score of detection box," in 2024 IEEE/CVF Winter Conference on Applications of Computer Vision (WACV), 2024, pp. 6583-6592.

\bibitem{vukas2024inboostrack}
Vukasin D. Stanojevic and Branimir T. Todorovic, "BoostTrack: Boosting the similarity measure and detection confidence for improved multiple object tracking," Machine Vision and Applications, vol. 35, no. 53, 2024.

\bibitem{isard2001bramble}
M. Isard and J. MacCormick, "BraMBLe: A Bayesian multiple-blob tracker," in Proceedings Eighth IEEE International Conference on Computer Vision (ICCV 2001), 2001, pp. 34-41.

\bibitem{zia2005MCMC}
Zia Khan and T. Balch and F. Dellaert, "MCMC-based particle filter for tracking a variable number of interacting targets," IEEE Transactions on Pattern Analysis and Machine Intelligence, vol. 27, no .11, pp. 1805-1819, 2005.

\bibitem{michael2009robust}
Michael D. Breitenstein and Fabian Reichlin and Bastian Leibe and Esther Koller-Meier and Luc Van Gool, "Robust tracking-by-detection using a detector confidence particle filter," in 2009 IEEE 12th International Conference on Computer Vision (ICCV), 2009, pp. 1515-1522.

\bibitem{ming2009detection}
Ming Yang and Gengjun Lv and Wei Xu and Yihong Gong, "Detection driven adaptive multi-cue integration for multiple human tracking," in 2009 IEEE 12th International Conference on Computer Vision (ICCV), 2009, pp. 1554-1561.

\bibitem{kenji2024boosted}
Kenji Okuma and Ali Taleghani and Nando de Freitas and James J. Little and David G. Lowe, "A boosted particle filter: multitarget detection and tracking," in Proceedings of European Conference on Computer Vision (ECCV), 2024, pp.28-39.

\bibitem{toan2025pami}
Toan Van Nguyen and Rasmus G. K. Christiansen and Dirk Kraft and Leon Bodenhagen, "GenTrack: A New Generation of Multi-Object Tracking," arXiv, 2025. Available at: \href{https://doi.org/10.48550/arXiv.2510.24399}{\url{https://doi.org/10.48550/arXiv.2510.24399}}.

\bibitem{toan2025icra}
Toan Van Nguyen and Rasmus G. K. Christiansen and Dirk Kraft and Leon Bodenhagen, "GenTrack2: An Improved Hybrid Approach for Multi-Object Tracking, arXiv, 2025. Available at: \href{https://doi.org/10.48550/arXiv.2510.24410}{\url{https://doi.org/10.48550/arXiv.2510.24410}}.

\bibitem{kennedy1995particle}
J. Kennedy and R. Eberhart, "Particle swarm optimization," in Proceedings of IEEE International Conference on Neural Networks, 1995, pp. 1942-1948.

\bibitem{clerc2002PSO}
M. Clerc and J. Kennedy, "The particle swarm - explosion, stability, and convergence in a multidimensional complex space," IEEE Transactions on Evolutionary Computation, vol. 6, no. 1, pp. 58-73, 2002.

\bibitem{kuhn1955hungarian}
H. W. Kuhn, "The Hungarian method for the assignment problem," Naval Research Logistics Quarterly, vol. 2, no. 1-2, pp. 83-97, 1995.

\bibitem{tingpeng2016improved}
Tingpeng Li and Yue Li and Yanling Qian, "Improved Hungarian algorithm for assignment problems of serial-parallel systems," Journal of Systems Engineering and Electronics, vol. 27, no. 4, pp. 858-870, 2016.

\bibitem{ayokor2007dynamics}
Ayorkor Mills-Tettey and Anthony Stent and M. Bernardine Dias, "The Dynamic Hungarian Algorithm for the Assignment Problem with Changing Costs," Robotics Institute, Carnegie Mellon University, 2007.

\bibitem{dendorfer2021motchallenge}
Patrick Dendorfer and Aljosa Osep and Anton Milan and Konrad Schindler and Daniel Cremers and Ian Reid and Stefan Roth and Laura Leal-Taixe, "MOTChallenge: A benchmark for single-camera multiple target tracking," International Journal of Computer Vision, vol. 129, pp. 845-881, 2021.

\bibitem{pattrick2020mot20}
Patrick Dendorfer and Hamid Rezatofighi and Anton Milan and Javen Shi and Daniel Cremers and Ian Reid and Stefan Roth and Konrad Schindler and Laura Leal-Taixe, "MOT20: A benchmark for multi object tracking in crowded scenes," arXiv, 2020. Available at: \href{https://doi.org/10.48550/arXiv.2003.09003}{\url{https://doi.org/10.48550/arXiv.2003.09003}}

\bibitem{vasant2006performance}
Vasant Manohar and Padmanabhan Soundararajan and Harish Raju and Dmitry Goldgof and Rangachar Kasturi and John Garofolo, "Performance evaluation of object detection and tracking in video, "Springer, Berlin, Heidelberg, 2006, in Narayanan, P.J., Nayar, S.K., Shum, HY. (eds) Computer Vision – ACCV 2006. ACCV 2006. Lecture Notes in Computer Science, vol .3852, pp. 151-161.

\bibitem{luiten2021hota}
Jonathon Luiten and Aljosa Osep and Patrick Dendorfer and Philip Torr and Andreas Geiger and Laura Leal-Taixe and Bastian Leibe, "HOTA: A higher order metric for evaluating multi-object tracking," International Journal of Computer Vision, vol. 129,
pp. 548-578, 2021.

\end{thebibliography}


\end{document}